\pdfoutput=1
\documentclass[twocolumn]{bmcart}

\usepackage{etoolbox}
\makeatletter
\patchcmd\set@numberlines@box{\rlap}{\@gobble}{}{}
\makeatother

\usepackage{amsthm, amssymb, amsmath, bm, graphicx, multirow, url, stfloats, etoolbox, boldline, rotating, array, cleveref, ragged2e}

\RequirePackage[numbers]{natbib}
\usepackage[utf8]{inputenc} 

\startlocaldefs
\endlocaldefs

\begin{document}

\begin{frontmatter}

\begin{fmbox}
\dochead{Research}


\title{Exploring QSAR Models for Activity-Cliff Prediction}


\author[
	addressref={aff1},                    
	email={markus.dablander@maths.ox.ac.uk}
]{\inits{M.F.D.}\fnm{Markus} \snm{Dablander}}
\author[
	addressref={aff3},
	email={Thierry.Hanser@lhasalimited.org}
]{\inits{T.H.}\fnm{Thierry} \snm{Hanser}}
\author[
	addressref={aff1},
	email={renaud.lambiotte@maths.ox.ac.uk}
]{\inits{R.L.}\fnm{Renaud} \snm{Lambiotte}}
\author[
	addressref={aff2},
	corref={aff2},
	email={garrett.morris@dtc.ox.ac.uk}
]{\inits{G.M.M.}\fnm{Garrett M.} \snm{Morris}}


\address[id=aff1]{%
	\orgdiv{Mathematical Institute},
	\orgname{University of Oxford},
	\street{Andrew Wiles Building, Radcliffe Observatory Quarter (550), Woodstock Road},
	\postcode{OX2 6GG,}
	\city{Oxford},
	\cny{United Kingdom}
}

\address[id=aff3]{%
	\orgname{Lhasa Limited},
	\street{Granary Wharf House, 2 Canal Wharf},
	\postcode{LS11 5PS,}
	\city{Leeds},
	\cny{United Kingdom}
}

\address[id=aff2]{%
	\orgdiv{Department of Statistics},
	\orgname{University of Oxford},
	\street{24-29 St Giles’},
	\postcode{OX1 3LB,}
	\city{Oxford},
	\cny{United Kingdom}
}





\begin{abstractbox}

\begin{abstract} 
\justifying
\parttitle{Introduction and Methodology} Pairs of similar compounds that only differ by a small structural modification but exhibit a large difference in their binding affinity for a given target are known as activity cliffs~(ACs). It has been hypothesised that QSAR models struggle to predict ACs and that ACs thus form a major source of prediction error. However, a study to explore the AC-prediction power of modern QSAR methods and its relationship to general QSAR-prediction performance is lacking. We systematically construct nine distinct QSAR models by combining three molecular representation methods (extended-connectivity fingerprints, physicochemical-descriptor vectors and graph isomorphism networks) with three regression techniques (random forests, k-nearest neighbours and multilayer perceptrons); we then use each resulting model to classify pairs of similar compounds as ACs or non-ACs and to predict the activities of individual molecules in three case studies: dopamine receptor D2, factor Xa, and SARS-CoV-2 main protease.

\parttitle{Results and Conclusions} We observe low AC-sensitivity amongst the tested models when the activities of both compounds are unknown, but a substantial increase in AC-sensitivity when the actual activity of one of the compounds is given. Graph isomorphism features are found to be competitive with or superior to classical molecular representations for AC-classification and can thus be employed as baseline AC-prediction models or simple compound-optimisation tools. For general QSAR-prediction, however, extended-connectivity fingerprints still consistently deliver the best performance. Our results provide strong support for the hypothesis that indeed QSAR methods frequently fail to predict ACs. We propose twin-network training for deep learning models as a potential future pathway to increase AC-sensitivity and thus overall QSAR performance.

\end{abstract}




\centering
\includegraphics[width=1.946\linewidth]{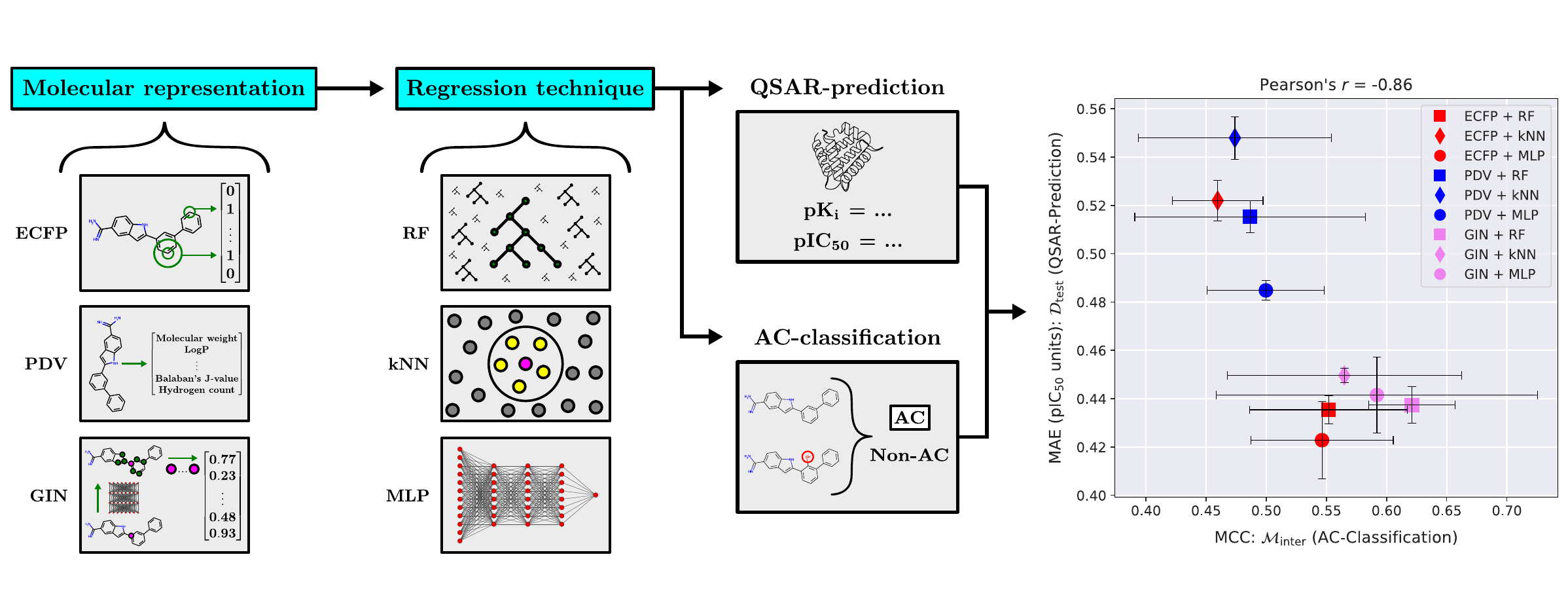}

\end{abstractbox}

\end{fmbox}

\end{frontmatter}




\text{} \newpage

\section*{Introduction}
\vspace{6pt}

\begin{figure*}[hb]
	\centering
	\includegraphics[width=1.85\linewidth]{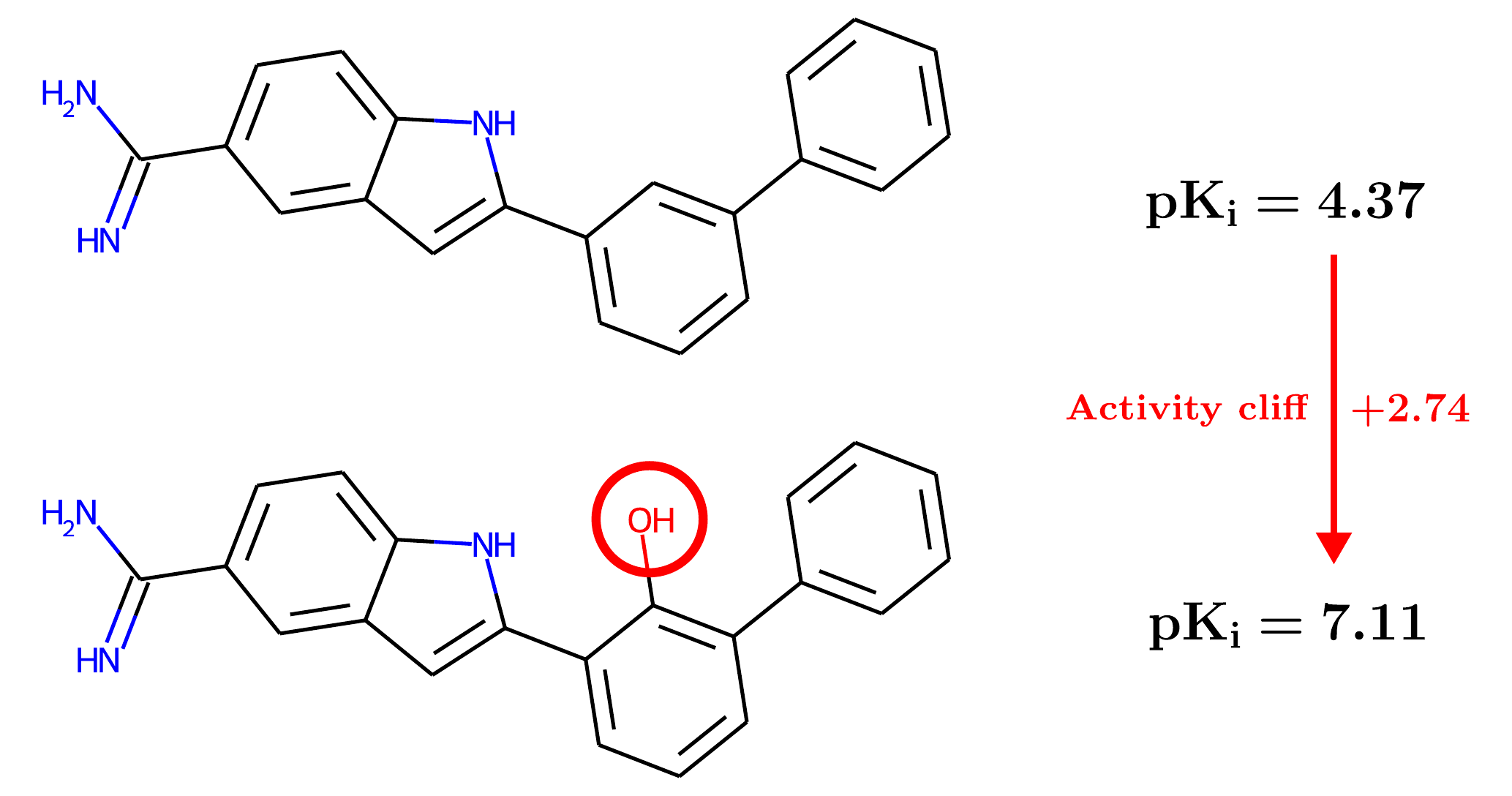}
	
	\caption{ Example of an activity cliff (AC) for blood coagulation factor Xa. A small structural transformation in the upper compound leads to an increase in inhibitory activity of almost three orders of magnitude. Both compounds were identified in the same ChEMBL assay with ID 658338.}
	\label{fig:ac_example_factor_Xa_CHEMBL658338}
\end{figure*}

Activity cliffs~(ACs) are pairs of small molecules that exhibit high structural similarity but at the same time show an unexpectedly large difference in their binding affinity against a given pharmacological target~\cite{silipo1991qsar, maggiora_outliers_2006, sheridan_experimental_2020, cruz-monteagudo_activity_2014, stumpfe_recent_2014, stumpfe_evolving_2019, stumpfe_advances_2020}. The existence of ACs directly defies the intuitive idea that chemical compounds with similar structures should have similar activities, often referred to as the \textit{molecular similarity principle}. An example of an AC between two inhibitors of blood coagulation factor Xa~\cite{leadley2001coagulation} is depicted in \Cref{fig:ac_example_factor_Xa_CHEMBL658338}; a small chemical modification involving the addition of a hydroxyl group leads to an increase in inhibition of almost three orders of magnitude.

For medicinal chemists, ACs can be puzzling and confound their understanding of structure-activity relationships (SARs)~\cite{stumpfe_recent_2014, vogt2011activity, dimova_activity_2015}. ACs reveal small compound-modifications with large biological impact and thus represent rich sources of pharmacological information. Mechanisms by which a small structural transformation can give rise to an AC include a drastic change in 3D-conformation and/or the switching to a different binding mode or even binding site. ACs form discontinuities in the SAR-landscape and can therefore have a crucial impact on the success of lead-optimisation programmes. While knowledge about ACs can be powerful when trying to escape from flat regions of the SAR-landscape, their presence can be detrimental in later stages of the drug development process, when multiple molecular properties beyond mere activity need to be balanced carefully to arrive at a safe and effective compound~\cite{cruz-monteagudo_activity_2014, stumpfe_recent_2014}. 

In the field of computational chemistry, ACs are suspected to form one of the major roadblocks for successful quantitative structure-activity relationship~(QSAR) modelling~\cite{golbraikh2014data, cruz-monteagudo_activity_2014, maggiora_outliers_2006, sheridan_experimental_2020}; abrupt changes in potency are expected to negatively influence machine learning algorithms for pharmacological activity prediction. During the development of QSAR models, ACs are sometimes dismissed as measurement errors~\cite{medina2013activity}, but simply removing ACs from a training data set can result in a loss of precious SAR-information~\cite{cruz2016probing}. 

Golbraikh~et~al.~\cite{golbraikh2014data} developed the MODI metric to quantify the smoothness of the SAR-landscape of binary molecular classification data sets and showed that the SAR-landscape smoothness is a strong determinant for downstream QSAR-modelling performance. In a related work, Sheridan~et~al.~\cite{sheridan_experimental_2020} found that the density of ACs in a molecular data set is strongly predictive of its overall modelability by classical descriptor- and fingerprint-based QSAR methods. Furthermore, they found that such methods incur a significant drop in performance when the test set is restricted to ``cliffy" compounds that form a large number of ACs. In a more extensive study, van Tilborg~et~al.~\cite{van2022exposing} observed a similar drop in performance when testing classical and graph-based QSAR techniques on compounds involved in ACs. Notably, in both studies this performance drop was also observed for highly nonlinear and adaptive deep learning models. In fact, van Tilborg reports that descriptor-based QSAR methods even outperform more complex deep learning models on ``cliffy" compounds associated with ACs. This runs counter to earlier hopes expressed in the literature that the approximation power of deep neural networks might ameliorate the problem of ACs~\cite{winkler2017performance}.

While these works provide valuable insights into the detrimental effects of SAR discontinuity on QSAR models, they consider ACs mainly indirectly by focussing on \textit{individual} compounds involved in ACs. Arguably, a distinct and more natural approach would be to investigate ACs directly at the level of compound \textit{pairs}. This approach has been followed in the AC-prediction field which is concerned with developing techniques to classify whether a pair of similar compounds forms an AC or not. An effective AC-prediction method would be of high value for drug development with important applications in rational compound optimisation and automatic SAR-knowledge acquisition. 

The AC-prediction literature is still very thin compared to the QSAR-prediction literature. An attempt to conduct an exhaustive literature review on AC-prediction techniques revealed a total number of $15$ methods~\cite{heikamp_prediction_2012, tamura_ligand-based_2020, de_la_vega_de_leon_prediction_2014, beck2014quantitative, namasivayam_searching_2012, namasivayam_prediction_2013, husby_structure-based_2015, horvath_prediction_2016, perez-benito_predicting_2019, asawa_prediction_2020, keyvanpour2021pcac, iqbal_prediction_2021, park2022acgcn, chen2022deepac}, all of which have been published since 2012. Current AC-prediction methods are often based on creative ways to extract features from pairs of molecular compounds in a manner suitable for standard machine learning pipelines. For example, Horvath et~al.~\cite{horvath_prediction_2016} used condensed graphs of reactions~\cite{hoonakker2011condensed,jauffret1990machine}, a representation technique originally introduced for modelling of chemical reactions, to encode pairs of similar compounds and subsequently predict ACs. Another method was recently described by Iqbal et~al.~\cite{iqbal_prediction_2021} who investigated the abilities of convolutional neural networks operating on 2D images of compound pairs to distinguish between ACs and non-ACs. Interestingly, none of the AC-prediction methods we identified employ feature extraction techniques built on modern graph neural networks (GNNs)~\cite{duvenaud2015convolutional, kipf2016semi, gilmer2017neural, velivckovic2017graph, xu2018powerful} with the exception of Park et al.~\cite{park2022acgcn} who recently applied graph convolutional methods to compound-pairs to predict ACs.

In spite of the existence of advanced AC-prediction models there are significant gaps left in the current AC-prediction literature. Note that any QSAR model can immediately be repurposed as an AC-prediction model by using it to individually predict the activities of two structurally similar compounds and then thresholding the predicted absolute activity difference. Nevertheless, at the moment there is no study that uses this straightforward technique to investigate the potential of current QSAR models to classify whether a pair of compounds forms an AC or not. Importantly, this also entails that the most salient AC-prediction models~\cite{heikamp_prediction_2012, de_la_vega_de_leon_prediction_2014, horvath_prediction_2016, tamura_ligand-based_2020, iqbal_prediction_2021} have not been compared to a simple QSAR-modelling baseline applied to compound pairs. It is thus an open question to what extent (if at all) these tailored AC-prediction techniques outcompete repurposed QSAR methods in the detection of ACs. This is especially relevant in light of the fact that several published AC-predict¸ion models~\cite{heikamp_prediction_2012, de_la_vega_de_leon_prediction_2014, iqbal_prediction_2021} are evaluated via compound-pair-based data splits which incur a significant overlap between training set and test set at the level of individual molecules; this type of data split should strongly favour standard QSAR models for AC-prediction, yet a comparison to such baseline methods is lacking.

We address these gaps by systematically investigating the abilities of nine frequently used QSAR models to classify pairs of similar compounds as ACs or non-ACs within three pharmacological data sets: dopamine receptor D2, factor Xa, and SARS-CoV-2 main protease. Each QSAR model is constructed by combining a molecular representation method (physicochemical-descriptor vectors~(PDVs)~\cite{todeschini2008handbook}, extended-connectivity fingerprints~(ECFPs)~\cite{rogers2010extended}, or graph isomorphism networks~(GINs)~\cite{xu2018powerful}) with a regression technique (random forests~(RFs), k-nearest neighbours~(kNNs), or multilayer perceptrons~(MLPs)). All models are used for two distinct prediction tasks: QSAR-prediction at the level of individual molecules, and AC-classification at the level of compound-pairs. The main contribution of this study is to shed light on the following questions:
\vspace{0.1pt}

\begin{itemize}
	
	\item What is the relationship between the ability of a QSAR model to predict the activities of individual compounds, versus its ability to classify whether pairs of similar compounds form ACs?
	
	\item When (if at all) are common QSAR models capable of predicting ACs?
	
	\item When (if at all) are common QSAR models capable of predicting which of two similar compounds is the more active one?
	
	\item Which QSAR model shows the strongest AC-prediction performance, and should thus be used as a baseline against which to compare tailored AC-prediction models?
	
	\item Do differentiable GINs outperform classical non-trainable ECFPs and PDVs as molecular representations for QSAR- and/or AC-prediction?
	
	\item How could ACs potentially be used to improve QSAR-modelling performance?
	
\end{itemize}

\section*{Experimental Methodology}
\vspace{6pt}

\subsection*{Molecular Data Sets}
\vspace{5pt}

We built three binding affinity data sets of small-molecule inhibitors of dopamine receptor D2, factor Xa, and SARS-CoV-2 main protease. Factor Xa is an enzyme in the coagulation cascade and a canonical target for blood-thinning drugs~\cite{leadley2001coagulation}. Dopamine receptor D2 is the main site of action for classic antipsychotic drugs which act as antagonists of the D2 receptor~\cite{seeman1987dopamine}. SARS-CoV-2 main protease is one of the key enzymes in the viral replication cycle of the SARS coronavirus 2, that recently caused the unprecedented COVID-19 pandemic; it is one of the most promising targets for antiviral drugs against this coronavirus~\cite{ullrich2020sars}.

For dopamine receptor D2 and factor Xa, data was extracted from the ChEMBL database~\cite{liu2007bindingdb} in the form of SMILES strings with associated K\textsubscript{i}~[nM] values. For SARS-CoV-2 main protease, data was obtained from the COVID moonshot project~\cite{achdout2020covid} in the form of SMILES strings with associated IC\textsubscript{50}~[µM] values. SMILES strings were standardised and desalted via the ChEMBL structure pipeline~\cite{bento2020open}. This step also removed solvents and all isotopic information. Following this, SMILES strings that produced error messages when turned into an RDKit mol object were deleted. Finally, a scan for duplicate molecules was performed: If the activities in a set of duplicate molecules were within the same order of magnitude then the set was unified via geometric averaging. Otherwise, the measurements were considered unreliable and the corresponding set of duplicate molecules was removed. This procedure reduced the data set for dopamine receptor D2 / factor Xa / SARS-CoV-2 main protease from $8883$ / $4116$ / $1926$ compounds to $6333$ / $3605$ / $1924$ unique compounds whereby $174$ / $21$ / $0$ sets of duplicate SMILES were removed and the rest was unified.

\subsection*{Activity Cliffs: Definition of Binary Classification Tasks}
\vspace{5pt}

The exact definition of an AC hinges on two concepts: structural similarity and large activity difference. An elegant technique to measure structural similarity in the context of AC analysis is given by the matched molecular pair~(MMP) formalism~\cite{kenny2005structure, hu_extending_2012}. An MMP is a pair of compounds that share a common structural core but differ by a small chemical transformation at a specific site. \Cref{fig:ac_example_factor_Xa_CHEMBL658338} depicts an example of an MMP whose variable parts are formed by a hydrogen atom and a hydroxyl group. To detect MMPs algorithmically, we used the \texttt{mmpdb} \texttt{Python}-package provided by Dalke et al.~\cite{dalke2018mmpdb}. We restricted ourselves to MMPs with the following commonly used~\cite{heikamp_prediction_2012, horvath_prediction_2016, tamura_ligand-based_2020} size constraints: the MMP core was required to contain at least twice as many heavy atoms as either of the two variable parts; each variable part was required to contain no more than $13$ heavy atoms; the maximal size difference between both variable parts was set to eight heavy atoms; and bond cutting was restricted to single exocyclic bonds. To guarantee a well-defined mapping from each MMP to a unique structural core, we canonically chose the core that contained the largest number of heavy atoms whenever there was ambiguity.
\begin{table*}[t]
	\centering
	\normalsize
	{\renewcommand{\arraystretch}{1.3}
		\begin{tabular}{p{3.75cm} V{4} p{3.75cm}|p{3.75cm}|p{3.75cm}}

			\textbf{Data Set} & 
			\textbf{Dopamine Receptor D2} & 
			\textbf{Factor Xa} & 
			\textbf{SARS-CoV-2 \newline Main Protease} 
			\\ \hlineB{4}
			
			\textbf{Compounds} & 
			$6333$ & 
			$3605$ & 
			$1924$
			\\ \hline
			
			\textbf{MMPs} & 
			$35484$ & 
			$21292$ & 
			$12594$
			\\ \hline
			
			\textbf{ACs} & 
			$461$ & 
			$1896$ & 
			$521$
			\\ \hline
			
			\textbf{Half-ACs} & 
			$3804$ & 
			$4693$ & 
			$1762$
			\\ \hline
			
			\textbf{Non-ACs} & 
			$31219$ & 
			$14703$ & 
			$10311$
			\\ \hline
			
			\textbf{ACs : Non-ACs} & 
			$\approx 1 : 68$ & 
			$\approx 1 : 8$ & 
			$\approx 1: 20$
			\\
	\end{tabular}}

	\vspace*{6mm}
	
	\caption[Sizes of data sets for QSAR/AC-prediction study.]{Sizes of our curated data sets and their respective numbers of matched molecular pairs (MMPs), activity cliffs (ACs), half-activity-cliffs (half-ACs) and non-activity-cliffs (non-ACs).}
	
	\label{tab: qsar_ac_study_datsets_overview}
\end{table*}
Based on the ratio of the activity values of both MMP compounds, each MMP was assigned to one of three classes: ``AC", ``non-AC" or ``half-AC". In accordance with the literature~\cite{heikamp_prediction_2012, namasivayam_prediction_2013, horvath_prediction_2016, vogt2011activity, bajorath_exploring_2014} we assigned an MMP to the ``AC"-class if both activity values differed by at least a factor of $100$. If both activity values differed by no more than a factor of $10$, then the MMP was assigned to the ``non-AC"-class. In the residual case the MMP was assigned to the ``half-AC"-class. To arrive at a well-separated binary classification task, we labelled all ACs as positives and all non-ACs as negatives. The half-ACs were removed and not considered further in our experiments. It is relevant to know the direction of a potential activity cliff, i.e.~which of the compounds in the pair is the more active one. We thus assigned a binary label to each MMP indicating its potency direction~(PD). PD-classification is a balanced binary classification task. \Cref{tab: qsar_ac_study_datsets_overview} gives an overview of all our curated data sets.

\subsection*{Data Splitting Technique} \label{subsec_data_splitting}
\vspace{5pt}

ACs are molecular pairs rather than single molecules; it is thus not obvious how best to split up a chemical data set into non-overlapping training- and test sets for the fair evaluation of an AC-prediction method. There seems to be no consensus about which data splitting strategy should be canonically used. Several authors~\cite{heikamp_prediction_2012, de_la_vega_de_leon_prediction_2014, iqbal_prediction_2021} have employed a random split at the level of compound pairs. While this technique is conceptually straightforward, it must be expected to incur a significant overlap between training- and test set at the level of individual molecules. For example, randomly splitting up a set of three MMPs $\{\{s_1, s_2\}, \{s_1, s_3\}, \{s_2, s_3\}\}$ into a training- and a test set might lead to $\{s_1, s_2\}$ and $\{s_1, s_3\}$ getting assigned to the training- and $\{s_2,s_3\}$ getting assigned to the test set which leads to a full inclusion of the test set in the training set at the level of individual molecules. This molecular overlap is problematic for at least three reasons: Firstly, it likely leads to overly optimistic results for AC-prediction methods since they will have already encountered some of the test compounds during training. Secondly, it does not model the natural situation encountered by medicinal chemists who we assume will not know the activity value of at least one compound in a test-set pair. Thirdly, the mentioned molecular overlap should lead to strong AC-prediction results for standard QSAR models, but to the best of our knowledge, no such control experiments have been run in the literature.

Horvath et al.~\cite{horvath_prediction_2016} and Tamura et al.~\cite{tamura_ligand-based_2020} have made efforts to address the shortcomings of a compound-pair-based random split. They came up with advanced data splitting algorithms designed to mitigate the molecular-overlap problem by either managing distinct types of test sets according to compound membership in the training set or by designing splitting techniques based on the structural cores of MMPs. However, their data splitting schemes exhibit a relatively high degree of complexity which can make their implementation and interpretation difficult. 

We propose a novel data splitting method which represents a favourable trade-off between rigour, interpretability and simplicity. Our technique shares some of its concepts with the methods proposed by Horvath et al.~\cite{horvath_prediction_2016} and Tamura et al.~\cite{tamura_ligand-based_2020} but might be simpler to implement and interpret. We first split the data into a training- and test set at the level of individual molecules and then use this basic split to distinguish several types of test sets at the level of compound pairs. Let
$$\mathcal{D} = \{s_1, s_2,...\}$$
be the given data set of individual molecules. Furthermore, let 
$$\mathcal{M} \subseteq \{ \{s, \tilde{s}\} \ \vert \ s \neq \tilde{s} \ \text{and} \ s, \tilde{s} \in \mathcal{D}	\} $$
be the set of all MMPs in $\mathcal{D}$ that have been labelled as either ACs or non-ACs. Each MMP $\{s, \tilde{s}\} \in \mathcal{M}$ shares a common structural core denoted as $\text{core}(\{s, \tilde{s}\})$. We use a random split to partition $\mathcal{D}$ into a training set $\mathcal{D}_{\text{train}}$ and a test set $\mathcal{D}_{\text{test}}$ and then define the following MMP-sets:
\begin{align*}
\mathcal{M}_{\text{train}} &= \{ \{s, \tilde{s}\} \in \mathcal{M} \ \vert \ s, \tilde{s} \in \mathcal{D}_{\text{train}}	\},	\\
\mathcal{M}^{}_{\text{inter}} &= \{ \{s, \tilde{s}\} \in \mathcal{M} \ \vert \ s \in \mathcal{D}_{\text{train}}, \ \tilde{s} \in \mathcal{D}_{\text{test}}	\} , \\
\mathcal{M}_{\text{test}} &= \{ \{s, \tilde{s}\} \in \mathcal{M} \ \vert \ s, \tilde{s} \in \mathcal{D}_{\text{test}}	\},\\
\mathcal{M}_{\text{cores}} &= \{ \{s, \tilde{s}\} \in \mathcal{M}_{\text{test}} \ \vert \ \text{core}(\{s, \tilde{s}\}) \notin \mathcal{C}_{\text{train}} 	\}.
\end{align*}
Here, $$\mathcal{C}_{\text{train}} = \{ \text{core}(\{s, \tilde{s}\}) \ \vert \ \{s, \tilde{s}\} \in \mathcal{M}_{\text{train}} \cup \mathcal{M}_{\text{inter}}	\}, $$
which describes the set of MMP-cores that appear in $\mathcal{D}_{\text{train}}$. 

\begin{figure*}
	\centering
	\includegraphics[width=2\linewidth]{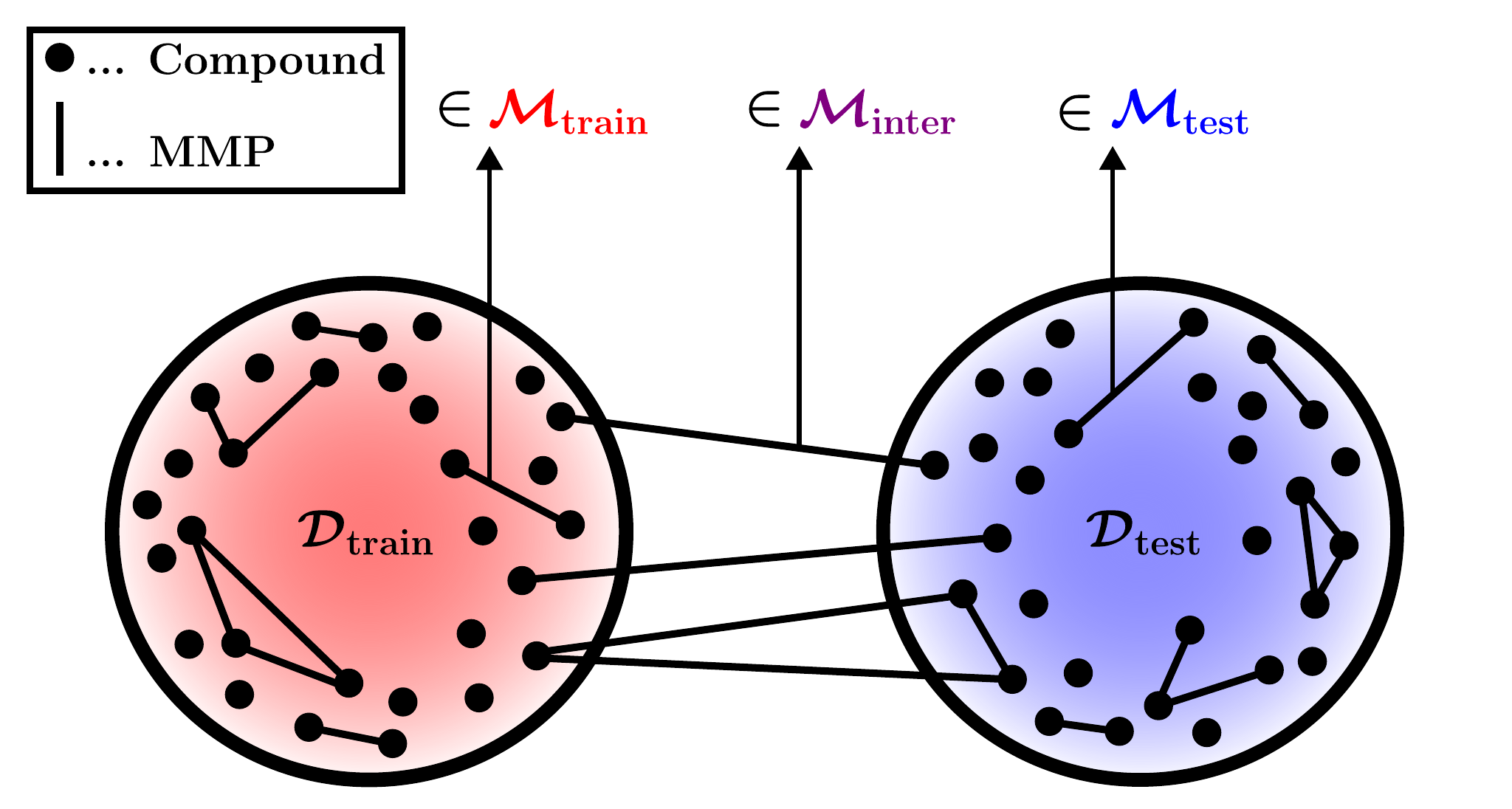}
	\caption{Illustration of our data splitting strategy. We distinguish between three MMP-sets, $\mathcal{M}_{\text{train}}, \mathcal{M}_{\text{inter}}$ and $\mathcal{M}_{\text{test}}$, depending on whether both MMP-compounds are in $\mathcal{D}_{\text{train}}$, one MMP-compound is in $\mathcal{D}_{\text{train}}$ and the other one is in $\mathcal{D}_{\text{test}}$, or both MMP-compounds are in $\mathcal{D}_{\text{test}}$. We additionally consider a fourth MMP-set, $\mathcal{M}_{\text{cores}}$, consisting of the MMPs in $\mathcal{M}_{\text{test}}$ whose structural cores do not appear in $\mathcal{M}_{\text{train}} \cup \mathcal{M}_{\text{inter}}$. }
	\label{fig:data_splitting_strategy}
\end{figure*}

Note that $\mathcal{M}_{\text{train}} \cup \mathcal{M}_{\text{inter}} \cup \mathcal{M}_{\text{test}} = \mathcal{M}$. The pair $(\mathcal{D}_{\text{train}}, \mathcal{M}_{\text{train}})$ describes the training space at the level of individual molecules and MMPs, and can be used to train a QSAR- or AC-prediction method. A trained method can then classify MMPs in $\mathcal{M}_{\text{test}}$, $\mathcal{M}_{\text{inter}}$ and $\mathcal{M}_{\text{cores}}$. $\mathcal{M}_{\text{test}}$ models an AC-prediction setting where the activities of both MMP-compounds are unknown. $\mathcal{M}_{\text{cores}}$ represents the subset of MMPs in $\mathcal{M}_{\text{test}}$ whose structural cores do not appear in $\mathcal{M}_{\text{train}} \cup \mathcal{M}_{\text{inter}}$; $\mathcal{M}_{\text{cores}}$ thus models the difficult task of predicting ACs in a strucurally novel area of chemical space. Finally, $\mathcal{M}_{\text{inter}}$ represents an AC-prediction scenario where the activity of one MMP-compound is given \textit{a priori}; this can be interpreted as a compound-optimisation task where one strives to predict small AC-inducing modifications of a query compound with known activity. An illustration of our data splitting strategy is given in \Cref{fig:data_splitting_strategy}.

We implemented our data splitting strategy within a $k$-fold cross validation scheme repeated with $m$ random seeds. This generated data splits of the form 
$$\mathcal{S}^{ij} = (\mathcal{D}^{ij}_{\text{train}}, \mathcal{D}^{ij}_{\text{test}}, \mathcal{M}^{ij}_{\text{train}},\mathcal{M}^{ij}_{\text{test}}, \mathcal{M}^{ij}_{\text{inter}}, \mathcal{M}^{\text{ij}}_{\text{cores}}) $$
for $i \in \{1,...,m\}$ and $j \in \{1,...,k\}$ where $(\mathcal{D}^{ij}_{\text{train}}, \mathcal{D}^{ij}_{\text{test}})$ represents the $j$-th split of $\mathcal{D}$ in the cross validation round with random seed $i$. The overall QSAR- and AC-prediction performance of each model was recorded as the average over the $mk$ training- and test runs based on all data splits $\mathcal{S}^{1,1}, ..., \mathcal{S}^{mk}$. We chose the configuration $(k,m) = (2,3)$ which gave a good trade-off between computational costs and accuracy and reasonable numbers of MMPs in the compound-pair-sets. In particular, random cross-validation with $k = 2$ gave expected relative sizes of:
$$\vert \mathcal{M}_{\text{train}} \vert : \vert \mathcal{M}_{\text{inter}} \vert : \vert \mathcal{M}_{\text{test}} \vert = 1 : 2 : 1 .$$ 
On average, $12.7 \ \%$, $11.91 \ \%$, and $6.84 \ \%$ of MMPs in $\mathcal{M}_{\text{test}}$ were also in $\mathcal{M}_{\text{cores}}$ for dopamine receptor D2, factor Xa, and SARS-CoV-2 main protease, respectively.

\subsection*{Prediction Strategies and Performance Measures}
\vspace{5pt}

In a data split of the form
$$\mathcal{S} = (\mathcal{D}_{\text{train}}, \mathcal{D}_{\text{test}}, \mathcal{M}_{\text{train}},\mathcal{M}_{\text{test}}, \mathcal{M}_{\text{inter}}, \mathcal{M}_{\text{cores}}) $$
each individual compound,
$s \in \mathcal{D}_{\text{train}} \cup \mathcal{D}_{\text{test}} = \mathcal{D},$
can be associated with an activity label $\text{a}(s) \in \mathbb{R}$, defined as the negative decadic logarithm of the experimentally measured activity of~$s$. We stuck with the canonical units used in the ChEMBL database and the COVID moonshot project ([nM] for K\textsubscript{i} and [µM] for IC\textsubscript{50}); each activity label $a(s)$ thus represents a standard pK\textsubscript{i}- or pIC\textsubscript{50} value (with an additive shift towards $0$ caused by the units which might slightly benefit prediction techniques initialised around the origin).
We are interested in QSAR-prediction functions,
$$ f : \mathcal{D} \to \mathbb{R},$$
that can map a chemical structure $s \in \mathcal{D}$ to an estimate of its binding affinity $a(s)$. The mapping $f$ is found via an algorithmic training process on the labelled data set 
$$\{ (s, a(s))	\ \vert \ s \in \mathcal{D}_{\text{train}} \}$$
and can then either be used to predict the activity labels of compounds in $\mathcal{D}_{\text{test}}$, or it can be repurposed to classify whether an MMP forms an activity cliff (AC-classification) and what the potency direction of an MMP is (PD-classification). If $\{s, \tilde{s}\} \in \mathcal{M}_{\text{inter}}$, then one can assume that the activity label of one of the compounds, say $a(s)$, is known; $f$ is then used to classify $\{s, \tilde{s}\}$ via:
\begin{align*}
\{s, \tilde{s}\} \mapsto
\begin{cases}
	\text{Non-AC} \quad &\text{if} \ \vert a(s) - f(\tilde{s}) \vert \leq d_{\text{crit}}, \\
	\text{AC} \quad &\text{if} \ \vert a(s) - f(\tilde{s}) \vert > d_{\text{crit}}.
\end{cases}
\end{align*}
Here $d_{\text{crit}} \in \mathbb{R}_{> 0}$ is a critical threshold above which an MMP is classified as an AC. Throughout this work we use $d_{\text{crit}} = 1.5$ (in pK\textsubscript{i}- or pIC\textsubscript{50} units) since this value represents the middle point between the intervals $[0, 1]$ and $[2, \infty)$ which correspond to absolute activity-label differences associated with non-ACs and ACs respectively.

If $\{s, \tilde{s}\} \in \mathcal{M}_{\text{test}} \cup \mathcal{M}_{\text{cores}}$ then the activities of both compounds are unknown and we classify $\{s, \tilde{s}\}$ via:
\begin{align*}
\{s, \tilde{s}\} \mapsto
\begin{cases}
\text{Non-AC} \quad &\text{if} \ \vert f(s) - f(\tilde{s}) \vert \leq d_{\text{crit}}, \\
\text{AC} \quad &\text{if} \ \vert f(s) - f(\tilde{s}) \vert > d_{\text{crit}}.
\end{cases}
\end{align*}
PD-classification for MMPs is performed in a straightforward manner: the activity labels of both MMP-compounds are predicted via $f$ and then compared to classify which compound is the more active one. 

The performance of $f$ for standard QSAR prediction in $\mathcal{D}_{\text{test}}$ is measured via the mean absolute error (MAE). For the balanced PD-classification problem we rely on accuracy as a suitable performance measure. For the highly imbalanced task of AC-classification, however, we use the Matthews correlation coefficient~(MCC), as well as sensitivity and precision. For the relatively small SARS-CoV-2 main protease data set we sometimes encountered the edge case where there were no positive predictions; we then set $\text{MCC} = 0$ and ignored ill-defined precision measurements when averaging the performance metrics to obtain the final results.

\subsection*{Molecular Representation- and Regression Techniques}
\vspace{5pt}

We constructed nine QSAR models via a robust combinatorial methodology that systematically combines three molecular representation methods with three regression techniques. This setup allows, for example, to compare the performance of molecular representation methods across regression techniques, data sets and predictions tasks.

For molecular representation, we used extended-connectivity fingerprints~\cite{rogers2010extended} (ECFPs), physicochemical molecular descriptor vectors~\cite{todeschini2008handbook} (PDVs), and graph isomorphism networks (GINs)~\cite{xu2018powerful}. Both ECFPs and PDVs were computed via RDKit~\cite{landrum2006rdkit}. The ECFPs were chosen to use a radius of two, a length of $2048$ bits, and active chirality flags. The PDVs had a dimensionality of $200$ and were constructed using the general list of descriptors from the work of Fabian et al.~\cite{fabian2020molecular}. This list encompasses properties related to druglikeness, logP, molecular refractivity, electrotopological state, molecular graph-structure, fragment profile, charge, and topological surface properties. The GIN was implemented using PyTorch Geometric~\cite{fey2019fast} and consisted of a variable number of graph convolutional layers, each with two internal hidden layers with ReLU activations and batch normalisation~\cite{ioffe2015batch}. We further chose the maxpool operator which computes the component-wise maximum over all atom feature vectors in the final graph layer to obtain a graph-level representation.

Each molecular representation was used as an input featurisation for three regression techniques: random forests (RFs), k-nearest neigbours (kNNs) and multilayer perceptrons (MLPs). The RF- and kNN-models were implemented via scikit-learn~\cite{pedregosa2011scikit} and the MLP-models via PyTorch~\cite{paszke2019pytorch}. The MLPs used ReLU activations and batch normalisation at each hidden layer. 

\begin{figure*}
	\centering
	\includegraphics[width=2\linewidth]{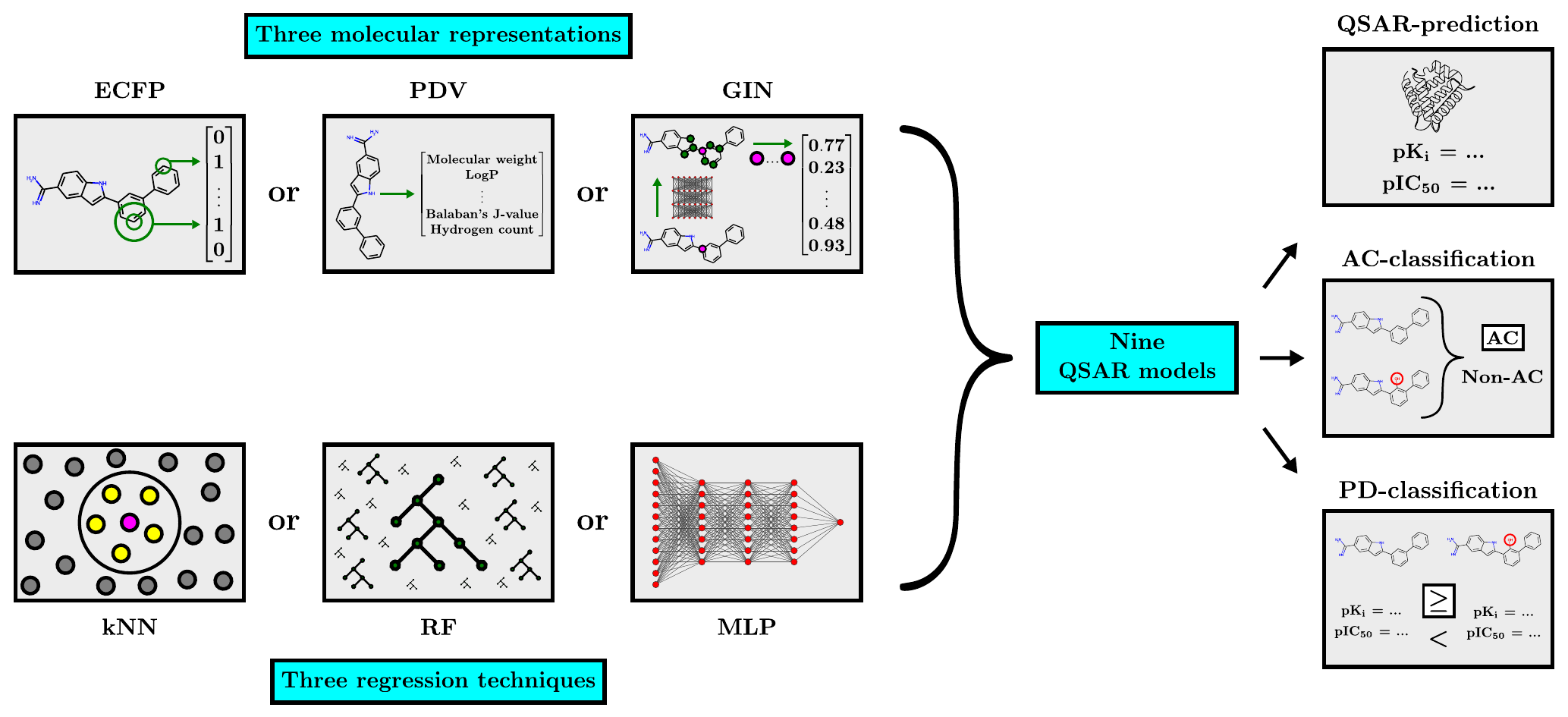}
	\caption{Schematic showing the combinatorial experimental methodology used for the study. Each molecular representation method is systematically combined with each regression technique, giving a total of nine QSAR models. Each QSAR model is trained and evaluated for QSAR-prediction, AC-classification and PD-classification within a $2$-fold cross validation scheme repeated with $3$ random seeds. For each of the $2*3 = 6$ trials, an extensive inner hyperparameter-optimisation loop on the training set is performed for each QSAR model.}
	\label{fig:methods_overview_linear_with_ac}
\end{figure*}

The GIN was combined with the regression techniques as follows: For MLP regression, the GIN was trained with the MLP as a projection head after the pooling step in the usual end-to-end manner. For RF- or kNN-regression, the GIN was first trained with a single linear layer added after the global pooling step that directly mapped the graph-level representation to an activity prediction. After this training phase the weights of the GIN were frozen and it was used as a static feature extractor. The RF- or kNN-regressor was then trained on the features extracted by the frozen GIN. \Cref{fig:methods_overview_linear_with_ac} illustrates our combinatorial experimental methodology.

\subsection*{Model Training and Hyperparameter Optimisation}
\vspace{5pt}

All models were trained using full inner hyperparame\-ter-optimisation loops. Hyperparameters of RFs and kNNs were optimised in scikit-learn~\cite{pedregosa2011scikit} by uniformly random sampling of hyperparameters from a predefined grid. The hyperparameters of MLPs and GINs were sampled from a predefined grid via the tree-structured Parzen estimator algorithm implemented in Optuna~\cite{akiba2019optuna}. Deep learning models were trained for $500$ epochs on a single NVIDIA GeForce RTX 3060 GPU via the mean squared error loss function using AdamW optimisation~\cite{loshchilov2017decoupled}. Weight decay, learning rate decay and dropout~\cite{srivastava2014dropout} were employed at all hidden layers for regularisation. Batch size, learning rate, learning rate decay rate, weight decay rate, and dropout rate were treated as hyperparameters and subsequently optimised. Note that the training length (i.e.~the number of gradient updates) was implicitly optimised by tuning the batch size for the fixed number of $500$ training epochs. Further implementation details can be found in our public code repository\footnote{\url{https://github.com/MarkusFerdinandDablander/QSAR-activity-cliff-experiments}}.

\section*{Results and Discussion}
\vspace{6pt}

\begin{figure*}[h!]
	\centering
	\includegraphics[width=2\linewidth]{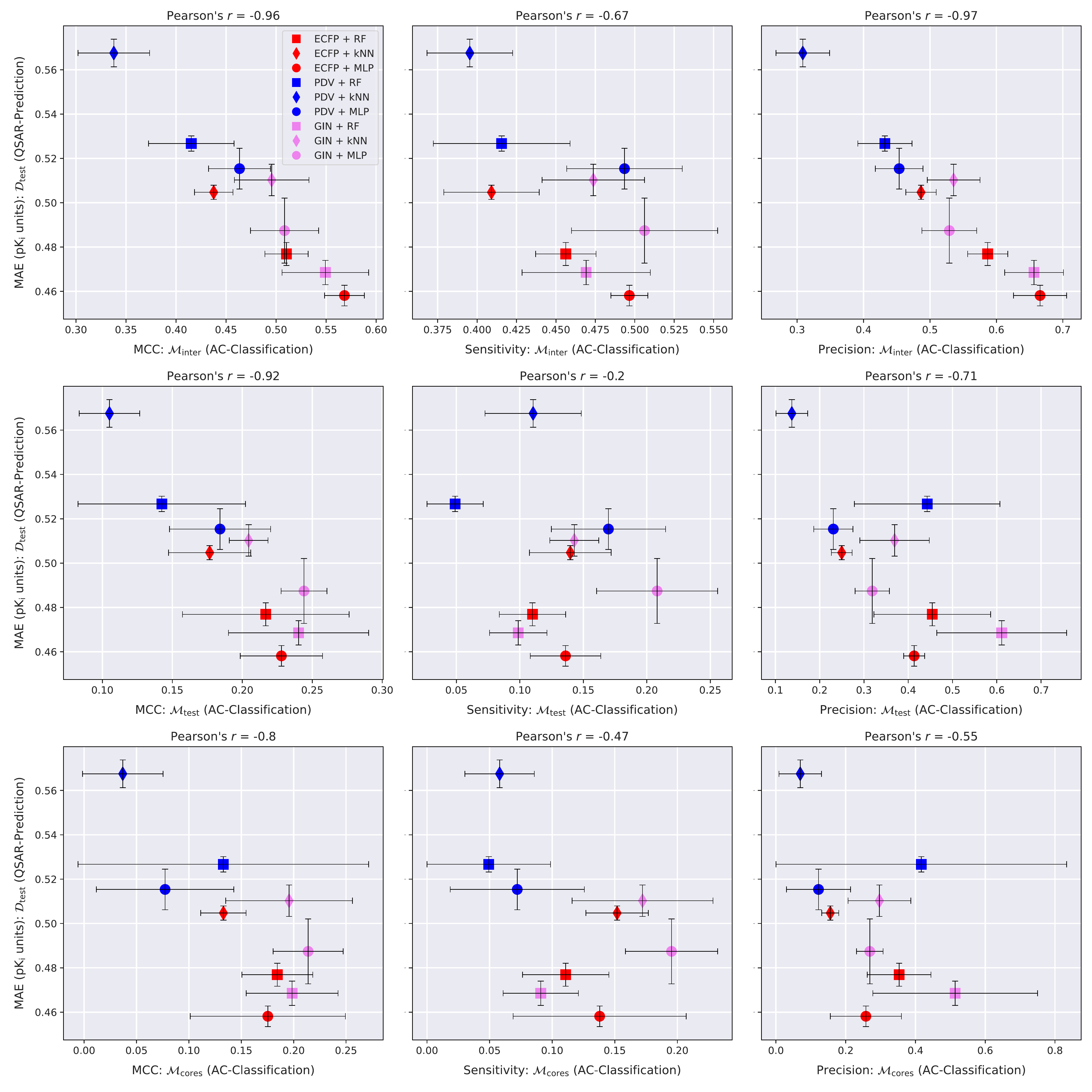}
	\caption{QSAR-prediction- and AC-classification results for \textbf{dopamine receptor D2}. For each plot, the $x$-axis corresponds to a combination of MMP-set and AC-classification performance metric and the $y$-axis shows the QSAR-prediction performance on the molecular test set $\mathcal{D}_{\text{test}}$. The total length of each error bar equals twice the standard deviation of the performance metric measured over all $mk = 3*2 = 6$ hyperparameter-optimised models. For each plot, the lower right corner corresponds to strong performance at both prediction tasks.}
	\label{fig:acpredresultschembldopamined2}
\end{figure*}
\begin{figure*}[h!]
	\centering
	\includegraphics[width=2\linewidth]{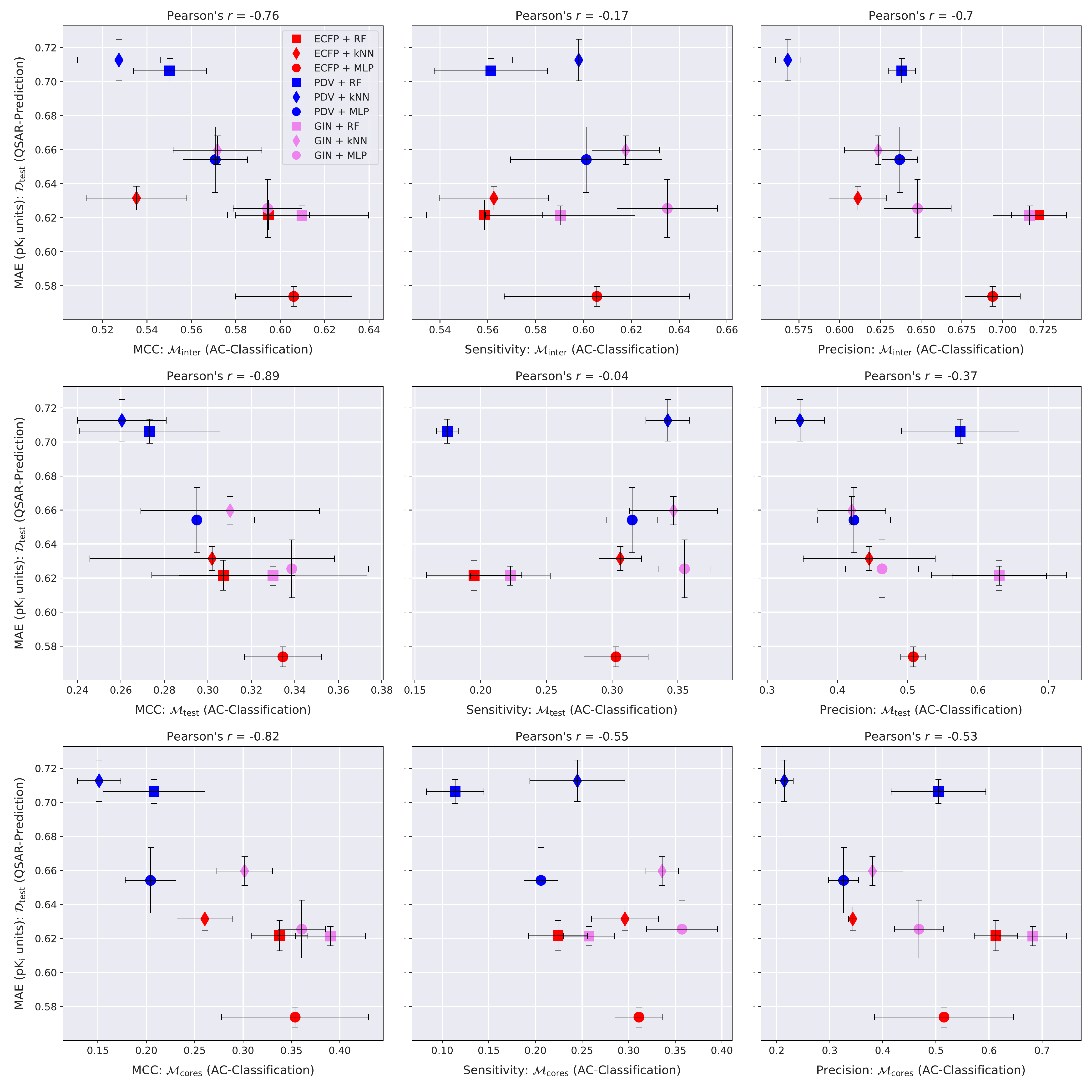}
	\caption{QSAR-prediction- and AC-classification results for \textbf{factor Xa}. For each plot, the $x$-axis corresponds to a combination of MMP-set and AC-classification performance metric and the $y$-axis shows the QSAR-prediction performance on the molecular test set $\mathcal{D}_{\text{test}}$. The total length of each error bar equals twice the standard deviation of the performance metric measured over all $mk = 3*2 = 6$ hyperparameter-optimised models. For each plot, the lower right corner corresponds to strong performance at both prediction tasks.}
	\label{fig:acpredresultschemblfactorxa}
\end{figure*}
\begin{figure*}[h!]
	\centering
	\includegraphics[width=2\linewidth]{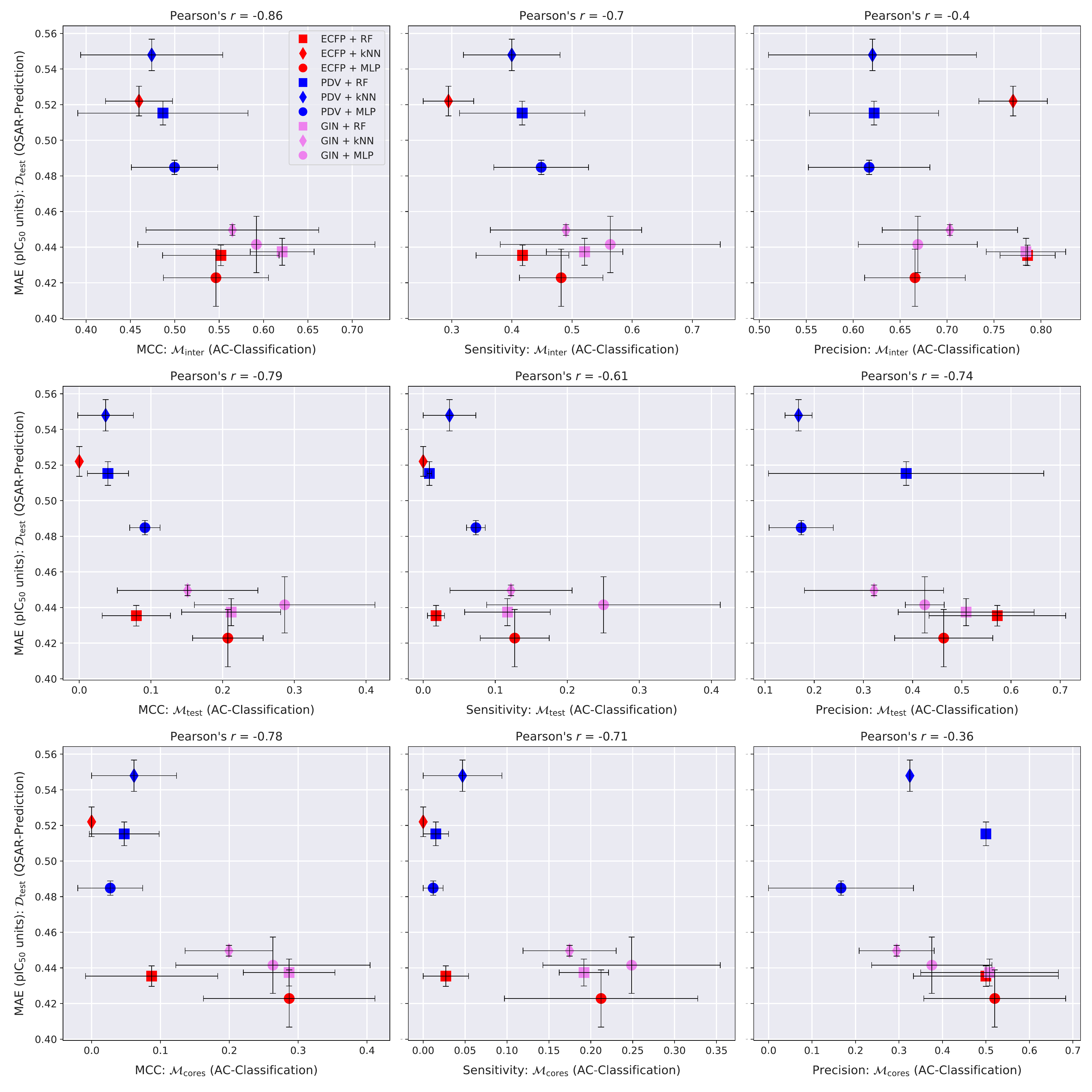}
	\caption{QSAR-prediction- and AC-classification results for \textbf{SARS CoV-2 main protease}. For each plot, the $x$-axis corresponds to a combination of MMP-set and AC-classification performance metric and the $y$-axis shows the QSAR-prediction performance on the molecular test set $\mathcal{D}_{\text{test}}$. The total length of each error bar equals twice the standard deviation of the performance metric measured over all $mk = 3*2 = 6$ hyperparameter-optimised models. The precision of the AC-classification task is lacking for the ECFP + kNN technique on $\mathcal{M}_{\text{test}}$ and $\mathcal{M}_{\text{cores}}$ since this method produced only negative AC-predictions for all trials on this data set. For each plot, the lower right corner corresponds to strong performance at both prediction tasks.}
	\label{fig:acpredresultsposterasarscov2mpro}
\end{figure*}
\begin{figure*}[h!]
	\centering
	\includegraphics[width=2\linewidth]{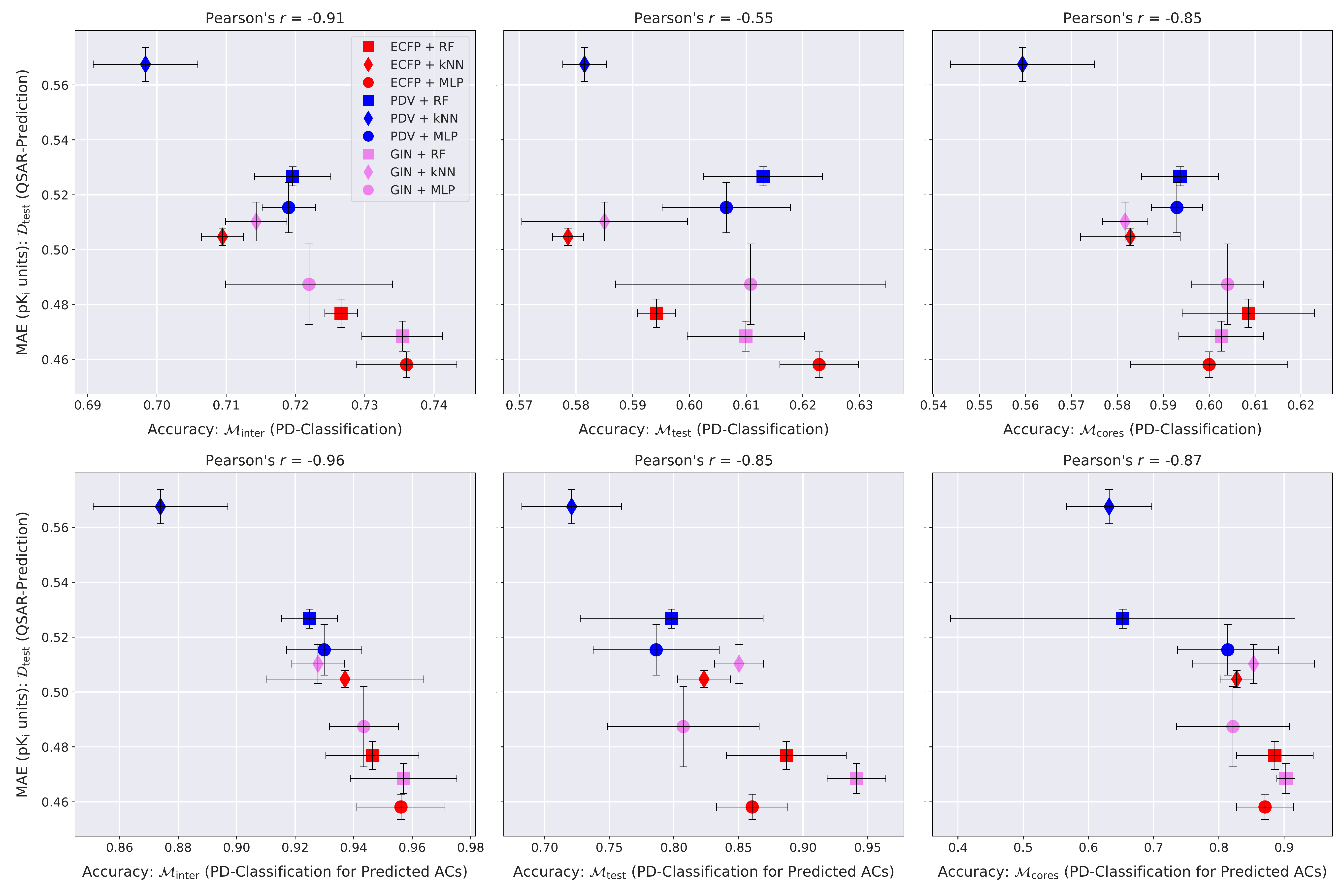}
	\caption{QSAR-prediction- and PD-classification results for \textbf{dopamine receptor D2}. Each column corresponds to an upper plot and a lower plot for one of the MMP-sets $\mathcal{M}_{\text{inter}}$, $\mathcal{M}_{\text{test}}$ or $\mathcal{M}_{\text{cores}}$. The x-axis of each upper plot indicates the PD-classification accuracy on the full MMP-set; the x-axis of each lower plot indicates the PD-classification accuracy on a restricted MMP-set only consisting of MMP predicted to be ACs by the respective method. The $y$-axis of each plot shows the QSAR-prediction performance on the molecular test set $\mathcal{D}_{\text{test}}$. The total length of each error bar equals twice the standard deviation of the performance metrics measured over all $mk = 3*2 = 6$ hyperparameter-optimised models. For each plot, the lower right corner corresponds to strong performance at both prediction tasks.}
	\label{fig:pdpredresultschembldopamined2}
\end{figure*}
\begin{figure*}[h!]
	\centering
	\includegraphics[width=2\linewidth]{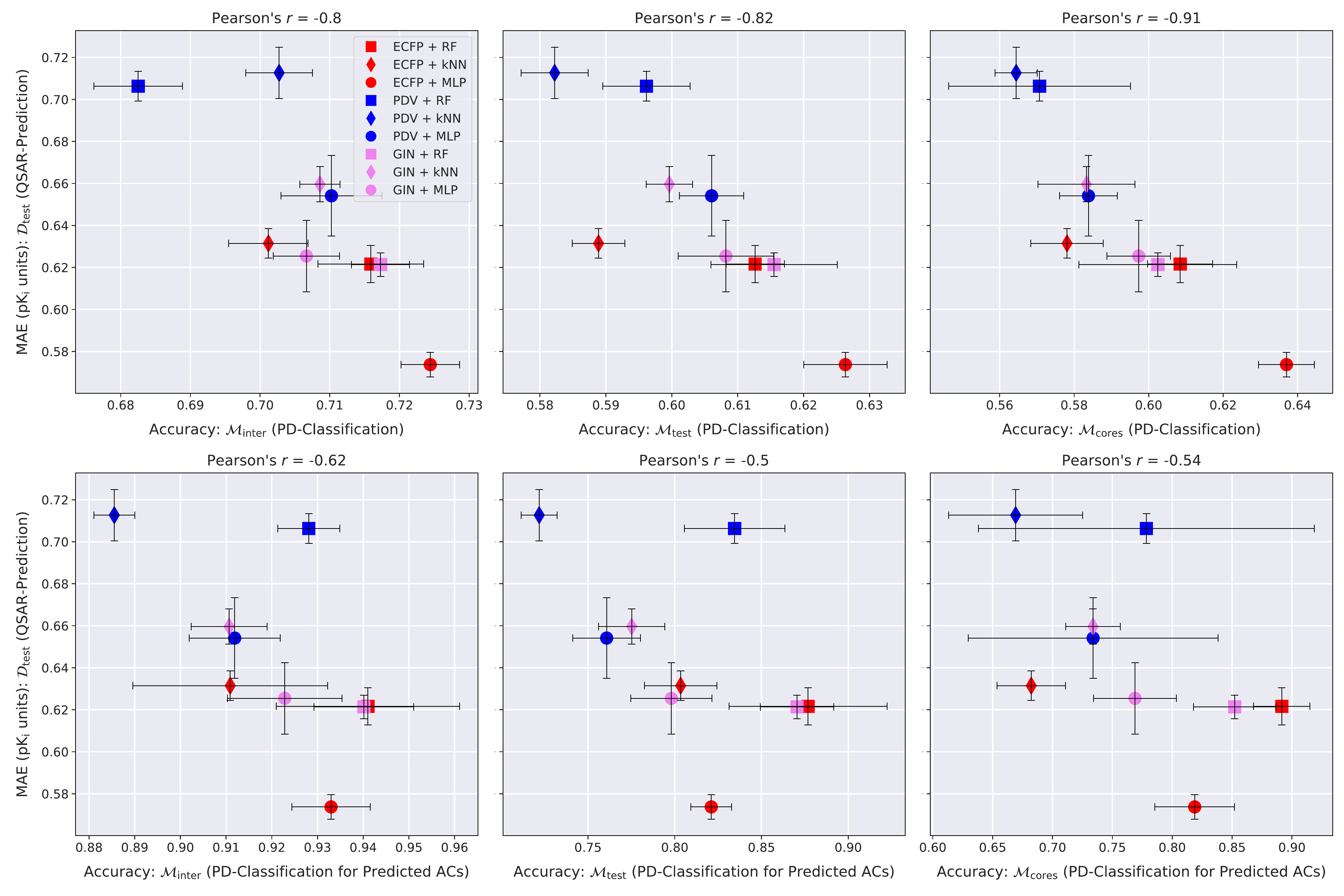}
	\caption{QSAR-prediction- and PD-classification results for \textbf{factor Xa}. Each column corresponds to an upper plot and a lower plot for one of the MMP-sets $\mathcal{M}_{\text{inter}}$, $\mathcal{M}_{\text{test}}$ or $\mathcal{M}_{\text{cores}}$. The x-axis of each upper plot indicates the PD-classification accuracy on the full MMP-set; the x-axis of each lower plot indicates the PD-classification accuracy on a restricted MMP-set only consisting of MMP predicted to be ACs by the respective method. The $y$-axis of each plot shows the QSAR-prediction performance on the molecular test set $\mathcal{D}_{\text{test}}$. The total length of each error bar equals twice the standard deviation of the performance metrics measured over all $mk = 3*2 = 6$ hyperparameter-optimised models. For each plot, the lower right corner corresponds to strong performance at both prediction tasks.}
	\label{fig:pdpredresultschemblfactorxa}
\end{figure*}
\begin{figure*}[h!]
	\centering
	\includegraphics[width=2\linewidth]{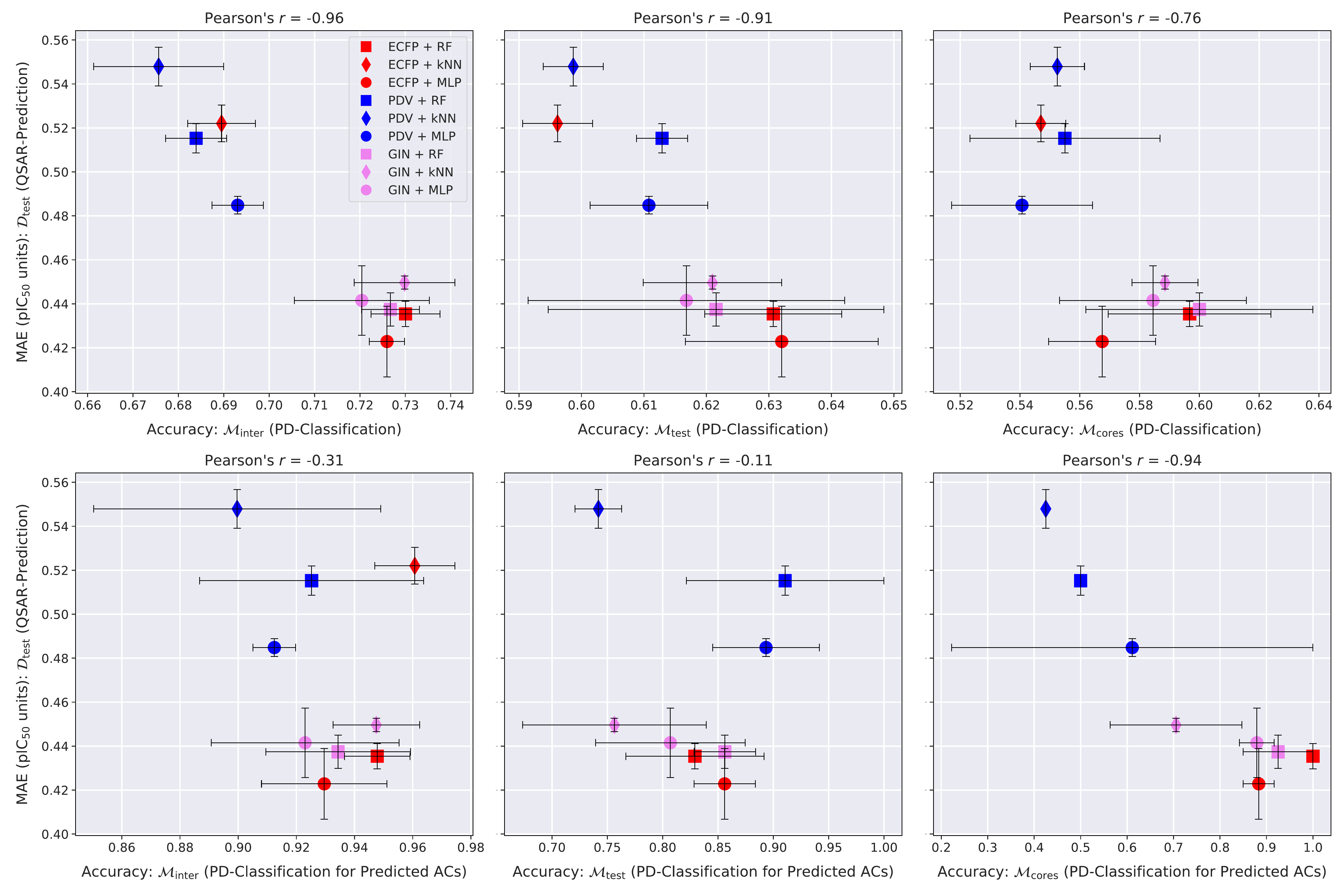}
	\caption{QSAR-prediction- and PD-classification results for \textbf{SARS-CoV-2 main protease}. Each column corresponds to an upper plot and a lower plot for one of the MMP-sets $\mathcal{M}_{\text{inter}}$, $\mathcal{M}_{\text{test}}$ or $\mathcal{M}_{\text{cores}}$. The x-axis of each upper plot indicates the PD-classification accuracy on the full MMP-set; the x-axis of each lower plot indicates the PD-classification accuracy on a restricted MMP-set only consisting of MMP predicted to be ACs by the respective method. The $y$-axis of each plot shows the QSAR-prediction performance on the molecular test set $\mathcal{D}_{\text{test}}$. The total length of each error bar equals twice the standard deviation of the performance metrics measured over all $mk = 3*2 = 6$ hyperparameter-optimised models. The accuracy of the PD-classification task for predicted ACs is lacking for the ECFP + kNN technique on $\mathcal{M}_{\text{test}}$ and $\mathcal{M}_{\text{cores}}$ since this method produced only negative AC-predictions for all trials on this data set. For each plot, the lower right corner corresponds to strong performance at both prediction tasks.}
	\label{fig:pdpredresultsposterasarscov2mpro}
\end{figure*}

The QSAR-prediction-, AC-classification- and PD-classification results for all three data sets are depicted in \Cref{fig:acpredresultschembldopamined2,fig:acpredresultschemblfactorxa,fig:acpredresultsposterasarscov2mpro,fig:pdpredresultschembldopamined2,fig:pdpredresultschemblfactorxa,fig:pdpredresultsposterasarscov2mpro}.

\subsection*{QSAR-Prediction Performance}
\vspace{5pt}

When considering the results depicted in~\Cref{fig:acpredresultschembldopamined2,fig:acpredresultschemblfactorxa,fig:acpredresultsposterasarscov2mpro,fig:pdpredresultschembldopamined2,fig:pdpredresultschemblfactorxa,fig:pdpredresultsposterasarscov2mpro} with respect to QSAR-prediction performance, one can see that ECFPs tend to lead to better performance (i.e.~a lower QSAR-MAE) compared to GINs, which in turn tend to lead to better performance compared to PDVs. In particular, the combination MLP-ECFP consistently produced the lowest QSAR-MAE across all three targets. These observations reinforce a growing corpus of literature that suggests that trainable GNNs have not yet reached a level of technical maturity by which they consistently and definitively outperform the much simpler non-differentiable ECFPs at important molecular property prediction tasks~\cite{stepivsnik2021comprehensive, mayr2018large, jiang2021could, menke2021using, chithrananda2020chemberta, sabando2021using, winter2019learning}.

\subsection*{AC-Classification Performance}
\vspace{5pt}

The AC-MCC plots in~\Cref{fig:acpredresultschembldopamined2,fig:acpredresultschemblfactorxa,fig:acpredresultsposterasarscov2mpro} reveal surprisingly strong overall AC-classification results on $\mathcal{M}_{\text{inter}}$. This type of MMP-set models a compound-optimisation scenario where a researcher strives to identify small structural modifications with a large impact on the activity of query compounds with known activities. For this task, a significant portion of our QSAR models exhibit an AC-MCC value greater than $0.5$ across targets, which appears impressive considering the simplicity of the approach. Exchanging $\mathcal{M}_{\text{inter}}$ with either $\mathcal{M}_{\text{test}}$ or $\mathcal{M}_{\text{cores}}$ leads to a substantial drop in the AC-MCC to approximately $0.3$ that appears to be mediated by a large drop in AC-sensitivity.

In most cases, GINs perform better than the other molecular representation methods with respect to the AC-MCC. Notably, kNN-regressors consistently perform best for AC-classification when combined with GIN-features; this supports the idea that GINs might have a heightened ability to resolve ACs by learning an embedding of chemical space in which the distance between two compounds is reflective of activity difference rather than structural difference. The combinations GIN-MLP, GIN-RF and ECFP-MLP exhibit particularly high AC-MCC values relative to the other methods. We recommend using at least one of these three models as a baseline against which to compare tailored AC-prediction models; the practical utility of any AC-prediction technique that cannot outperform these three common QSAR methods is questionable.

Across all three targets, AC-sensitivity is moderately high on $\mathcal{M}_{\text{inter}}$ but universally low on $\mathcal{M}_{\text{test}}$ and $\mathcal{M}_{\text{cores}}$. This is consistent with the hypothesis that ACs form one of the major sources of prediction error for QSAR models. The weak AC-sensitivity on $\mathcal{M}_{\text{test}}$ and $\mathcal{M}_{\text{cores}}$ indicates that modern QSAR methods are largely blind to ACs in novel areas of chemical space and thus lack essential chemical knowledge. GINs clearly outperform the other two more classical molecular representations across regression techniques with respect to AC-sensitivity. In particular, the GIN-MLP combination leads to the highest AC-sensitivity in all examined cases and thus discovers the most ACs. The highly parametric nature of GINs that makes them prone to overfitting could at the same time enable them to better model jagged regions of the SAR-landscape that contain ACs than classical task-agnostic representations.

There is a wide gap between distinct prediction techniques with respect to AC-precision: some models achieve a considerable level of AC-precision such that over $50$\% of positively predicted MMPs in $\mathcal{M}_{\text{test}}$ and $\mathcal{M}_{\text{cores}}$ are indeed actual ACs. Other QSAR models, however, seem to fail almost entirely with respect to this metric on $\mathcal{M}_{\text{test}}$ and $\mathcal{M}_{\text{cores}}$ and only deliver modest performance on $\mathcal{M}_{\text{inter}}$. RFs tend to exhibit the strongest AC-precision and the weakest AC-sensitivity. This might be as a result of their ensemble nature which should intuitively lead to conservative but trustworthy predictions of extreme effects such as ACs.

\subsection*{PD-Classification Performance}
\vspace{5pt}

The abilities of the evaluated QSAR models to identify which is the more active compound in an MMP is universally weak, with PD-accuracies clustering around $0.7$ on $\mathcal{M}_{\text{inter}}$ and around $0.6$ on $\mathcal{M}_{\text{test}}$ and $\mathcal{M}_{\text{cores}}$, as can be seen in the top rows of~\Cref{fig:pdpredresultschembldopamined2,fig:pdpredresultschemblfactorxa,fig:pdpredresultsposterasarscov2mpro}. Predicting the potency direction for two compounds with similar structures and thus usually similar activity levels must be considered a challenging task. The combination ECFP-MLP reaches the strongest PD-accuracy in the majority of cases and we recommend starting with this model as a baseline for more advanced PD-prediction methods.

One can argue that the activity order of two similar compounds is of little interest if the true activity difference is small, as is often the case. We therefore also restricted PD-classification to predicted ACs. The three plots in the bottom rows of~\Cref{fig:pdpredresultschembldopamined2,fig:pdpredresultschemblfactorxa,fig:pdpredresultsposterasarscov2mpro} depict the PD-accuracy of each QSAR model on the subset of MMPs that were also predicted to be ACs by the same model. In this practically more relevant scenario PD-prediction accuracy tends to exceed $0.9$ on $\mathcal{M}_{\text{inter}}$ and $0.8$ on $\mathcal{M}_{\text{test}}$ and $\mathcal{M}_{\text{cores}}$. The QSAR models investigated here are thus able to identify the correct activity order of MMPs if they also predict them to be ACs. The relatively rare instances in which the PD of a predicted AC is misclassified, however, reflect severe QSAR-prediction errors.

\subsection*{Linear Relationship between QSAR-MAE and AC-MCC}
\vspace{5pt}

Our experiments reveal a consistent linear relationship between the QSAR-MAE and the AC-MCC as can be seen in the left columns of~\Cref{fig:acpredresultschembldopamined2,fig:acpredresultschemblfactorxa,fig:acpredresultsposterasarscov2mpro}. A potential mechanism driving this effect could be that as the overall QSAR-MAE of a model improves, its accuracy at predicting activity differences between similar molecules might be expected to improve as well; previously misclassified MMPs whose predicted absolute activity differences were already close to the critical value $d_{\text{crit}} = 1.5$ might then gradually move to the correct side of the decision boundary and increase the AC-MCC. The results suggest that for real-world QSAR models the AC-MCC and the QSAR-MAE are strongly predictive of each other; while this observation only rests on nine models, it is highly consistent across MMP-sets and pharmacological targets.

\subsection*{Future Research: Exploring Twin-Network Training Schemes}
\vspace{5pt}

\begin{figure*}[h!]
	\centering
	\includegraphics[width=1.7\linewidth]{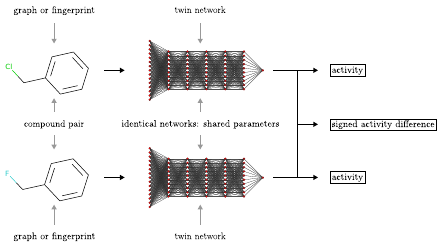}
	\caption{Twin-network training strategy for deep-learning-based QSAR models that might increase AC-sensitivity. Twin-network training could be conducted on general compound pairs and on MMPs, with larger weights given to MMPs associated with larger activity differences.}
	\label{fig:siamese_network}
\end{figure*} 

ACs are rich in pharmacological information; at the same time the experiments have shown that QSAR models exhibit low AC-sensitivity and thus frequently fail to predict ACs. In spite of this, to the best of our knowledge so far no method has been described to tackle this problem by attempting to increase the AC-sensitivity of QSAR models. We propose twin-network training of deep-learning models as a potential strategy to increase AC-sensitivity. Comparatively little work has been done to investigate twin neural network architectures (also referred to as \textit{Siamese} networks~\cite{chicco2021siamese, bromley1993signature, koch2015siamese, taigman2014deepface}) in computational drug discovery~\cite{dhami2019predicting, zhong2019graph, schwarz2020attentionddi, torres2020exploring,baskin2006neural, alvarez2010qt,chen2019multifaceted,fernandez2021siamese,jeon2019resimnet,roberts2019using,nourani2020tripletprot,gao2018interpretable}. However, twin networks provide a natural way to tackle chemical prediction problems on compound pairs such as AC-classification.

Instead of training a deep network, $f$, on an individual compound, $s$, with activity label, $a(s)$, via a classical squared error loss,
$(a(s) - f(s))^{2}$,
we suggest to train $f$ on compound \textit{pairs}, $\{s, \tilde{s}\}$, using a pair-based loss:
\begin{align*}
&w_{\{s, \tilde{s}\}}[ (a(s) - f(s))^{2} + (a(\tilde{s}) - f(\tilde{s}))^{2}
\\&+w_{\text{diff}} ((a(s) - a(\tilde{s})) - (f(s) - f(\tilde{s})))^{2}].
\end{align*}
The quantity $w_{\{s, \tilde{s}\}}$ is used to specify the weight put on the compound pair $\{s, \tilde{s}\}$ during training; $w_{\text{diff}}$ determines the relative importance of predicting the individual activities of $s$ and $\tilde{s}$ versus predicting the activity difference associated with $\{s, \tilde{s}\}$. Twin-network training could be conducted in two phases: first on general compound pairs in $\mathcal{D}_{\text{train}} \times \mathcal{D}_{\text{train}}$ and then on MMPs in $\mathcal{M}_{\text{train}}$. In the second phase, the weight function $w_{\{s, \tilde{s}\}}$ could be used to assign training weights to MMPs proportional to their associated activity differences; MMPs that represent larger activity differences might encode structural transformations that are pharmacologically more relevant and thus should receive more attention during training. This weighting procedure could lead to increased AC-sensitivity and the extraction of more chemical knowledge. Our pair-based training strategy is depicted in~\Cref{fig:siamese_network} and is based on a twin neural network model for AC-prediction with discrete outputs that we explored in a previous research study~\cite{dablander2021siamese}. We intend to evaluate the proposed twin-network training scheme in a future study.

\section*{Conclusions}
\vspace{6pt}

To the best of our knowledge this is the first study to investigate the AC-prediction capabilities of QSAR models. It is also the first work to explore the quantitative relationship between QSAR-prediction (at the level of individual molecules) and AC-prediction (at the level of compound-pairs). As part of our methodology we have additionally introduced a simple, interpretable, and rigorous data-splitting technique for pair-based prediction tasks.

When the activities of both MMP-compounds are unknown (i.e.~absent from the training set) then common QSAR models exhibit low AC-sensitivity which limits their utility for AC-prediction. This strongly supports the hypothesis that QSAR methods do indeed regularly fail to predict ACs which might thus form a major source of prediction errors in QSAR modelling~\cite{golbraikh2014data, cruz-monteagudo_activity_2014, maggiora_outliers_2006, sheridan_experimental_2020}. However, if the activity of one MMP-compound is known (i.e.~present in the training set) then AC-sensitivity increases substantially; for query compounds with known activities, QSAR methods can therefore be used as simple AC-prediction-, compound-optimisation- and SAR-knowledge-acquisition tools. Furthermore, based on the observed potency-directon (PD) classification results we can expect the estimated activity direction of predicted ACs to have a high degree of accuracy.

With respect to molecular representation, we have found robust evidence that non-trainable task-agnostic ECFPs still outcompete differentiable GINs at general QSAR-prediction. This adds to a growing awareness that standard message-passing GNNs might need to be improved further to definitively beat classical molecular featurisations such as ECFPs~\cite{stepivsnik2021comprehensive, mayr2018large, jiang2021could, menke2021using, chithrananda2020chemberta, sabando2021using, winter2019learning}. One potential angle to achieve this could be self-supervised GNN-pretraining, which has recently shown promising results in the molecular domain~\cite{hu2019strategies, wang2021molclr}. However, while GINs appear to be inferior to ECFPs for QSAR-prediction, they tend to be advantageous for AC-classification; their highly parametric nature might simultaneously lead to increased overfitting but to a better modelling of the more jagged regions of the SAR-landscape. We thus recommend using GINs as an AC-classification baseline since such an agreed-upon baseline is currently lacking.

Finally, the low AC-sensitivity of QSAR models when the activites of both MMP-compounds are unknown suggests that such methods are still lacking essential SAR knowledge; on the flip side, it might be possible to boost QSAR-modelling performance and increase the amount of extracted SAR knowledge by developing techniques to increase AC-sensitivity. To this end, we propose an AC-sensitive twin-network~\cite{chicco2021siamese, bromley1993signature, koch2015siamese, taigman2014deepface} training scheme for deep-learning models that we intend to explore in the future.


\begin{backmatter}
	
\section*{Funding} \justifying \noindent
This research was supported by the University of Oxford's UK EPSRC Centre For Doctoral Training in Industrially Focused Mathematical Modelling (EP/L015803/1) and by the not-for-profit organisation and educational charity Lhasa Limited (\url{https://www.lhasalimited.org/}).

\section*{Abbreviations}

\begin{itemize}
	\item AC = Activity Cliff
	\item ECFP = Extended-Connectivity Fingerprint
	\item GIN = Graph Isomorphism Network
	\item GNN = Graph Neural Network
	\item kNN = k-Nearest Neighbour
	\item MAE = Mean Absolute Error
	\item MCC = Matthews Correlation Coefficient
	\item MLP = Multilayer Perceptron
	\item MMP = Matched Molecular Pair
	\item PD = Potency Direction
	\item PDV = Physicochemical-Descriptor Vector
	\item QSAR = Quantitative Structure-Activity Relationship
	\item RF = Random Forest
	\item SAR = Structure-Activity Relationship
\end{itemize}

\section*{Availability of data and materials}

\noindent All used data sets, the code to reproduce and visualise the experimental results, and the exact numerical results generated by the original experiments are available in our public code repository \url{https://github.com/MarkusFerdinandDablander/QSAR-activity-cliff-experiments}.

\section*{Competing interests}
\noindent The authors declare that they have no competing interests.

\section*{Authors' contributions}

\noindent The computational study was designed, implemented, conducted and interpreted by the first author M.D.~The research was supervised by R.L., G.M.M., and T.H. who gave valuable scientific advice during weekly meetings. The computer code was written by M.D. The paper manuscript was written by M.D. Feedback was provided by R.L., G.M.M. and T.H. during the writing process. The novel data splitting technique for MMP-data, the QSAR-modelling-based activity cliff prediction strategies and the proposed twin-network training scheme were developed by M.D. All scientific figures were designed by M.D., with input from G.M.M., R.L. and T.H. All chemical data sets were gathered and cleaned by M.D. All authors read and approved the final manuscript.


\bibliographystyle{spbasic} 
\bibliography{bmc_article}      



\end{backmatter}

\end{document}